\documentclass[letterpaper]{article}
\usepackage[preprint]{aaai2027}
\usepackage[hyphens]{url}
\usepackage{graphicx}
\usepackage{natbib}
\usepackage{caption}
\usepackage{subcaption}
\usepackage{amsmath,amssymb,booktabs}
\usepackage{xcolor}
\newcommand{\affmark}[1]{\textsuperscript{\normalfont #1}}
\title{Beyond Future Views: Exploring Future-State World Models for \\ Vision-Language Navigation}
\author{%
  Kangyi Wu\affmark{1$\ast$}, Pengna Li\affmark{1}, Yucheng Wang\affmark{2$\dagger$}, Yance Fang\affmark{3}, Xihan Ma\affmark{2},\\
  Zhizhong Su\affmark{2}, Yang Wu\affmark{4}, Ye Deng\affmark{5}, Jinjun Wang\affmark{1}, Jianyi Liu\affmark{1$\ddagger$}%
}
\affiliations{\small
  \affmark{1}National Key Laboratory of Human-Machine Hybrid Augmented Intelligence,\\
  National Engineering Research Center for Visual Information and Applications,\\
  Institute of Artificial Intelligence and Robotics, Xi'an Jiaotong University, Xi'an, China\\
  \affmark{2}Horizon Robotics\\
  \affmark{3}School of Information, Renmin University of China, Beijing, China\\
  \affmark{4}School of Artificial Intelligence and Robotics, Xi'an University of Architecture and Technology, Xi'an, China\\
  \affmark{5}School of Computing and Artificial Intelligence, Southwestern University of Finance and Economics, Chengdu, China%
}

\begin{document}
\maketitle
\begingroup
\renewcommand{\thefootnote}{\fnsymbol{footnote}}
\footnotetext[1]{This work was done while Kangyi Wu was a Research Intern with Horizon Robotics.}
\footnotetext[2]{Project Leader.}
\footnotetext[3]{Corresponding Author. Email: \url{jyliu@xjtu.edu.cn}.}
\endgroup

\begin{abstract}
Vision-Language Navigation~(VLN) requires an embodied agent to follow language instructions through unseen 3D environments, where the target spaces often lie beyond the current observation. Navigation world models address this challenge by predicting future views either as pixels or latent representations, providing valuable look-ahead information for action decision. Both paradigms define the future primarily through what the agent may observe, raising a fundamental question: is predicting future views the ultimate form to represent the future for VLN? Cognitive studies show that the human hippocampus prospectively represents both future goals and intervening locations along planned routes. Inspired by this predictive-map view, we organize the navigation-relevant future around goal-directed spatial states and route structure. We instantiate this perspective with two complementary abstractions: a short-horizon waypoint that specifies the next local state to reach, and a mid-horizon corridor that captures the spatial support of the upcoming route. Based on this design, we propose FS-Nav, a future-state world modeling framework that predicts waypoint-corridor states from spatial memory and language instructions, and communicates them to a VLM navigator through top-down and first-person interfaces. Experiments on standard VLN benchmarks demonstrate that FS-Nav achieves state-of-the-art performance. This highlights navigation-relevant future-state modeling as a promising direction for representing the future in VLN.
\end{abstract}

\section{Introduction}
Vision-Language Navigation~(VLN)~\cite{anderson2018vision,ku2020room} requires an embodied agent to follow natural-language instructions and navigate to a target location in previously unseen 3D environments. Advances in vision-language models~(VLMs) have driven remarkable progress in VLN through their strong multimodal understanding and reasoning capabilities. Recent methods~\cite{wei2025streamvln,zhang2024navid,wu2026dual} therefore adopt VLMs as navigation backbones, directly mapping visual observations and language instructions to navigation actions. Despite these advances, directly predicting actions from current observations remains largely reactive~\cite{li2025regnav,lin2026bat}, without explicitly anticipating what may lie ahead.

\begin{figure}[t]
    \centering
    \includegraphics[width=\columnwidth]{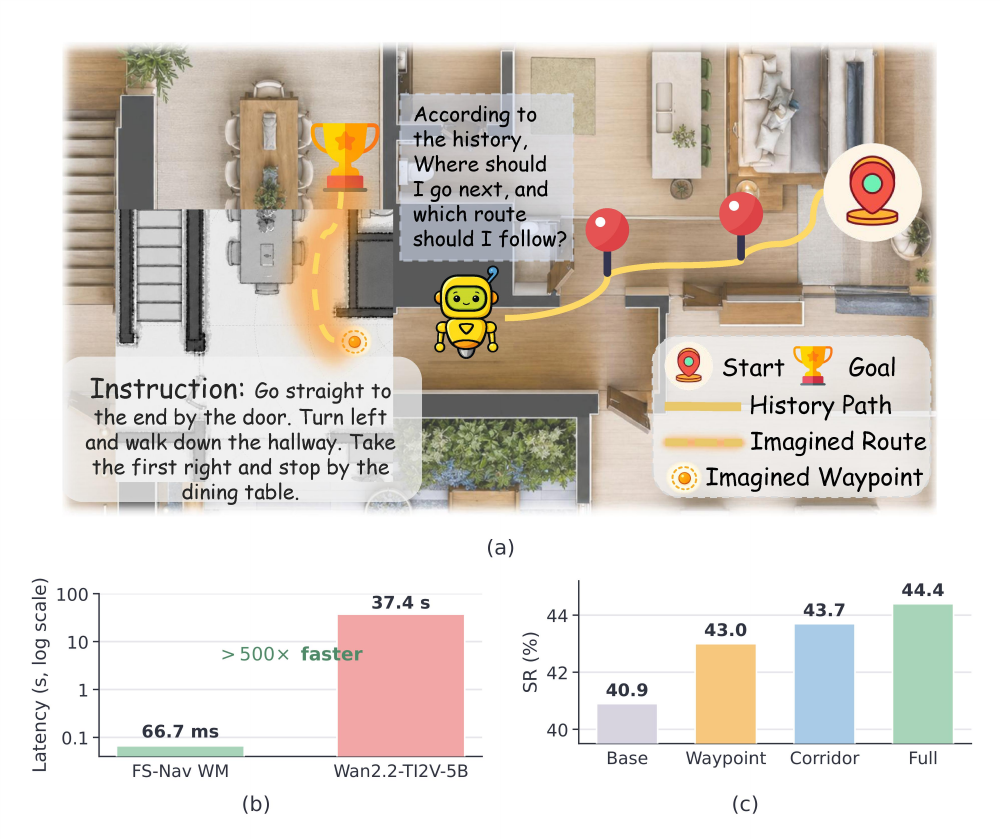}
      \caption{Motivation and key properties of future-state world model. (a) FS-Nav represents the upcoming route with a short-horizon waypoint and a mid-horizon corridor. (b) On a single RTX 4090, future-state prediction takes 66.7 ms, over $500\times$ faster than generating a five-frame future observation with Wan2.2-TI2V-5B. (c) The waypoint and corridor each improve navigation performance, with their combination achieving the highest SR on R2R-CE validation unseen.}
    \label{fig:motivation}
\end{figure}

Navigation world models address this limitation by predicting possible future outcomes before action selection. Early approaches synthesize future egocentric views at the pixel level, providing explicit look-ahead signals for navigation~\cite{bar2025navigation,liu2025navforeseeunifiedvisionlanguageworld}. Such fine-grained generation requires predicting every visual detail, which is difficult to anticipate and may be unnecessary for downstream decisions. More recent methods~\cite{yao2025navmorph,assran2025vjepa2selfsupervisedvideo} move beyond pixel reconstruction by predicting future visual representations in latent space. By abstracting away low-level appearance variations, they focus prediction on more stable, high-level visual content.
These advances make world models a powerful foundation for embodied intelligence. Despite this progression, both pixel- and latent-level approaches define the future through observations. We wonder whether
predicting what the agent will see is the ultimate form to represent the future in VLN.

A principled answer is suggested by predictive cognitive maps. Brown et al. found that the human hippocampus prospectively represents both future goals and intervening locations along the planned route during goal-directed navigation~\cite{brown2016prospective}. Predictive-map theory further describes a spatial state through the future states expected to be occupied under a behavior policy~\cite{stachenfeld2016hippocampus}. Together, these findings suggest that navigation can organize the future around goal-directed spatial states and route structure, rather than only future views.

Inspired by this perspective, we represent the navigation-relevant future at two complementary horizons: a short-horizon waypoint that specifies the next local state to reach, and a mid-horizon corridor that captures the spatial support of the upcoming route. Together, they form a compact, goal-conditioned abstraction of future navigation states, describing where the agent should move next and how the route should unfold, as illustrated in Figure~\ref{fig:motivation}(a).

Based on this representation, we propose FS-Nav, a future-state world modeling framework for VLN. FS-Nav consists of three components: a Future State World Model predicts waypoint-corridor states from the agent's accumulated spatial memory and language instruction; a Future State Renderer converts the predicted future state into two visual interfaces: a top-down canvas for explicit spatial planning and a first-person visual overlay for egocentric grounding. Finally, a Future-State-Guided Navigator consumes the resulting guidance together with visual observations and the instruction to predict navigation actions. This design separates what the world model predicts from how the prediction is visually grounded for navigation. As previewed in Figure~\ref{fig:motivation}(b) and (c), this compact representation is both efficient to predict and effective for navigation.

Our contributions are summarized as follows.
\begin{itemize}
    \item Drawing inspiration from predictive cognitive maps, we introduce a cognitively grounded perspective on navigation world modeling that organizes the anticipated future around goal-directed spatial states and route structure.
    \item We propose FS-Nav, which instantiates this perspective with a short-horizon waypoint and a mid-horizon corridor, predicts them from spatial memory and language instructions, and renders them through top-down and first-person interfaces to guide a VLM navigator.
    \item Extensive experiments on R2R-CE and RxR-CE demonstrate that FS-Nav achieves state-of-the-art performance, validating the effectiveness of navigation-relevant future-state modeling for VLN.
\end{itemize}

\section{Related Works}

\subsection{Vision-Language Navigation.}

Vision-Language Navigation~(VLN) requires an embodied agent to follow natural-language instructions in unseen environments~\cite{anderson2018vision,lyu2026instruction}. Earlier methods rely on modular perception and planning pipelines~\cite{li2026think} or select actions from predefined navigable candidates~\cite{an2024etpnav}. Recent advances in vision-language models~(VLMs) have enabled more unified navigation agents that directly map language instructions and visual histories to actions~\cite{wei2025streamvln,wei2025ground,li2026spaact}.
These methods improve embodied decision making through stronger temporal modeling, instruction grounding, and action prediction. However, they primarily reason from observations collected so far to determine what to do next, making navigation largely reactive. The spatial structure of the upcoming route is not explicitly modeled before action generation, but remains implicit within the VLM's reasoning process. In contrast, FS-Nav explicitly predicts the upcoming navigation space as a waypoint--corridor future state, providing the VLM navigator with predictive route guidance before action generation.
\subsection{Waypoint- and Route-Guided VLN.}

Continuous VLN methods commonly introduce intermediate spatial outputs to bridge instruction following and low-level control. Learned waypoint methods predict local navigable candidates or motion targets that are selected by a navigation policy and executed through a controller~\cite{krantz2021waypoint,hong2022bridging,an2024etpnav,lyu2026himemvln}. Recent approaches extend this formulation through image-grounded goals, short trajectories, or global route prediction. Goal2Pixel~\cite{bao2026goal2pixel} predicts a visible goal pixel that is back-projected into a waypoint and tracked by a local planner, while NavOne~\cite{zhan2026navone} predicts global path and goal distributions from top-down maps.
FS-Nav also predicts waypoint--corridor geometry, but assigns it a different role. Rather than treating the prediction as
the navigation decision or a directly executable plan, FS-Nav represents it as a multi-horizon future state that provides
predictive context to a closed-loop VLM navigator. This design makes the upcoming route explicit while preserving the VLM's
ability to adapt its actions to the instruction, real observations, and navigation history.

\begin{figure*}[t]
    \centering
    \includegraphics[width=0.9\textwidth]{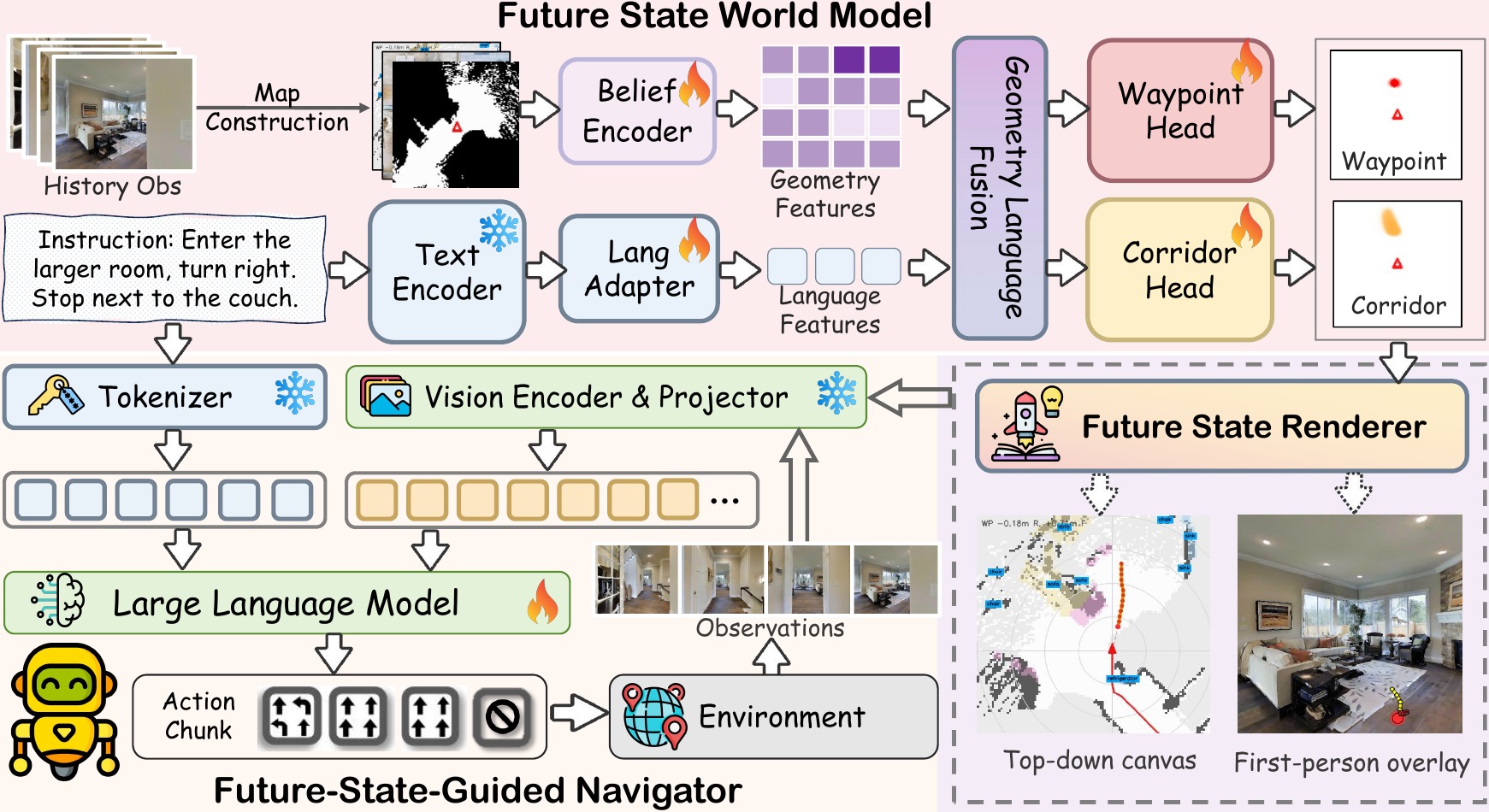}
    \caption{Overview of FS-Nav. The Future State World Model encodes the accumulated spatial memory and navigation instruction, fuses their geometry and language features, and predicts a short-horizon waypoint together with a mid-horizon corridor. The Future State Renderer deterministically renders the same prediction as either a top-down canvas or a first-person overlay. Conditioned on the instruction, observation history, and rendered guidance, the Future-State-Guided Navigator generates an action chunk and interacts with the environment in a closed loop. Flame symbols indicate trainable modules.}
    \label{fig:method}
\end{figure*}

\subsection{World Models in VLN.}

World models enable navigation agents to anticipate possible outcomes before acting.
Pixel-level world models explicitly synthesize future views. PathDreamer~\cite{koh2021pathdreamer} generates observations at unvisited viewpoints, while Navigation World Models~(NWM)~\cite{bar2025navigation} predict action-conditioned egocentric videos.
Although intuitive, pixel-level generation requires reconstructing high-dimensional appearance details that may be redundant or irrelevant to navigation. Latent-level world models reduce this burden by predicting future visual features or latent dynamics rather than raw RGB observations~\cite{yao2025navmorph,zhang2025latent,zhang2026rae,hao2026latentpilot}.
These representations provide a more compact prediction space, but retain an observation-centric formulation.

Recent studies have begun to explore more structured predictive representations.
SpaAct~\cite{li2026spaact} jointly models spatial transitions and actions, WorldVLN~\cite{zhao2026worldvln} predicts latent state transitions for aerial VLN, and WMNav~\cite{nie2025wmnav} maintains predictive state memories for object-goal navigation. These developments suggest that navigation world models need not be restricted to predicting future visual observations.
Following this broader predictive-state view, FS-Nav draws inspiration from predictive cognitive maps~\cite{brown2016prospective} and organizes the anticipated future around goal-directed spatial states and route structure. Its waypoint--corridor representation offers a compact and effective navigation-relevant predictive target.

\section{Method}
We propose FS-Nav, a future-state world modeling framework that makes the upcoming route explicit for VLN. As illustrated in Fig.~\ref{fig:method}, FS-Nav contains three components: a Future State World Model, a Future State Renderer, and a Future-State-Guided Navigator.

\subsection{Problem Formulation and Framework Overview}

We consider VLN in a continuous 3D environment.
At navigation step $t$, the agent receives a natural-language instruction~$\mathcal{L}$ and an egocentric observation $x_t$.
The observation history is denoted by
$O_{\leq t}=\{x_0,\ldots,x_t\}$.
Based on the observations collected so far, the agent constructs a spatial map $M_t$ that records occupancy, explored regions, and semantics.

Rather than predicting an unseen future observation, FS-Nav predicts a future navigation state
\begin{equation}
    \hat{S}_t
    =f_{\phi}(M_t,\mathcal{L})
    =\left(\hat{W}_t,\hat{C}_t\right),
\end{equation}
where $\hat{W}_t$ and $\hat{C}_t$ denote the predicted short-horizon waypoint and mid-horizon corridor, respectively. A deterministic renderer converts the same future state into a visual interface:
\begin{equation}
    Z_t^k
    =\mathcal{R}_k(\hat{S}_t,M_t,x_t),
    \qquad
    k\in\{\mathrm{canvas},\mathrm{fp}\}.
\end{equation}
The future-state guided navigator $p_{\theta}$ then predicts an action sequence~$A_t$ according to
\begin{equation}
    p_{\theta}
    \left(
    A_t\mid O_{\leq t},\mathcal{L},Z_t^k
    \right).
\end{equation}
In the following, we first define the waypoint--corridor future state and then detail the world model, renderer, and navigator.

\subsection{Navigation Future-State Representation}

\paragraph{Waypoint.}
The waypoint $W_t$ specifies the next local target.
We follow the remaining expert trajectory and select the first position whose cumulative path distance reaches a short-horizon distance $d_{\mathrm{w}}$; if the remaining trajectory is shorter, we use its endpoint.
After transforming this target into the agent-centered frame, we rasterize it as a Gaussian heatmap
$H_t^{W}\in[0,1]^{H\times W}$.
This provides smoother spatial supervision than directly regressing a single coordinate.
The predicted waypoint is recovered from the heatmap using differentiable soft-argmax~\cite{liu2025mind}.

\paragraph{Corridor.}
A waypoint indicates where to move next, but not how the route continues beyond it.
We therefore define a corridor $C_t$ using the subsequent future trajectory between the waypoint horizon $d_{\mathrm{w}}$ and a mid-horizon distance $d_{\mathrm{c}}$, subject to a maximum number of future steps or the end of the episode.
The trajectory is transformed into the agent-centered frame and rasterized as a soft path band
$H_t^{C}\in[0,1]^{H\times W}$.
We additionally retain the ordered points
$P_t^{C}=(p_{t,1},\ldots,p_{t,N})$
for direct path supervision.
We represent the corridor as $C_t=(H_t^C, P_t^C)$, where the path band captures its spatial support and the ordered points preserve route continuity and direction.

\subsection{Future State World Model}

The Future State World Model predicts $\hat{S}_t$ from the current spatial map~$M_t$ and instruction~$\mathcal{L}$. For the spatial map~$M_t$ constructed from historical observations, a belief encoder extracts the corresponding geometry feature. In parallel, a frozen Qwen2.5-VL~\cite{bai2025qwen2} text encoder produces language features, which are projected to the spatial feature dimension~\cite{lyu2026touchthinker}.
The geometry and language features are then fed into a Geometry--Language Fusion module. The resulting feature map $F_t$ jointly captures the observed environment geometry and the instruction-conditioned route information.
The fused feature map $F_t$ is subsequently passed to two state heads that predict the short-horizon waypoint and mid-horizon corridor, respectively.
The waypoint branch produces
\begin{equation}
    \hat{H}_t^{W}=\sigma\!\left(g_W(F_t)\right),
    \qquad
    \hat{W}_t=\operatorname{SoftArgmax}(\hat{H}_t^{W}).
\end{equation}
The corridor branch produces a route-band mask
\begin{equation}
    \hat{H}_t^{C}=\sigma\!\left(g_C(F_t)\right),
\end{equation}
and an ordered sequence of corridor points
\begin{equation}
    \hat{P}_t^{C}=g_P\!\left(\operatorname{Pool}(F_t)\right).
\end{equation}
Here, $\sigma$ denotes the sigmoid function, $g_W$ and $g_C$ are the waypoint and corridor-mask prediction heads, respectively, $\operatorname{Pool}(\cdot)$ denotes spatial pooling, and $g_P$ is the corridor-point regression head.
Together, they form the predicted corridor $\hat{C}_t=(\hat{H}_t^C, \hat{P}_t^C)$.
The mask supplies dense spatial supervision, while the ordered points preserve path direction and avoid relying only on mask-to-polyline post-processing.

We train the waypoint branch with heatmap reconstruction and coordinate regression~\cite{wu2025cem}:
\begin{equation}
    \mathcal{L}_{\mathrm{wp}}
    =\left\|\hat{H}_t^{W}-H_t^{W}\right\|_2^2
    +\lambda_{\mathrm{w}}
    \left\|\hat{W}_t-W_t\right\|_1.
\end{equation}
For the corridor mask, we combine pixel-wise binary cross-entropy, Dice loss, a centerline-aware clDice term, and total variation regularization:
\begin{equation}
\begin{split}
    \mathcal{L}_{\mathrm{mask}}
    ={}&\mathcal{L}_{\mathrm{BCE}}(\hat{H}_t^{C},H_t^{C})
    +\lambda_{\mathrm{dice}}\mathcal{L}_{\mathrm{Dice}}(\hat{H}_t^{C},H_t^{C})\\
    &+\lambda_{\mathrm{cl}}\mathcal{L}_{\mathrm{clDice}}(\hat{H}_t^{C},H_t^{C})
    +\lambda_{\mathrm{tv}}\mathcal{L}_{\mathrm{TV}}(\hat{H}_t^{C}).
\end{split}
\end{equation}
The ordered-point branch is supervised with a masked point-regression loss $\mathcal{L}_{\mathrm{pts}}$, while a curvature penalty discourages abrupt changes between consecutive path segments. The complete world-model objective is
\begin{equation}
\begin{split}
    \mathcal{L}_{\mathrm{wm}}
    ={}&\mathcal{L}_{\mathrm{wp}}
    +\mathcal{L}_{\mathrm{mask}}\\
    &{}+\lambda_{\mathrm{pts}}\mathcal{L}_{\mathrm{pts}}(\hat{P}_t^{C},P_t^{C})
    +\lambda_{\mathrm{curv}}\mathcal{L}_{\mathrm{curv}}(\hat{P}_t^{C}).
\end{split}
\end{equation}

\subsection{Future State Renderer}

The world model predicts route geometry, whereas the navigator operates on images and text. The Future State Renderer connects the two without predicting new information. It deterministically converts the same metric waypoint and corridor into one of two visual inputs for the navigator.

\paragraph{Top-down canvas.}
The canvas shows the current agent pose, accumulated local map, predicted waypoint, and predicted corridor in a bird's-eye view. It is appended to the observation sequence as an additional image. This view exposes relative position, free space, and route shape directly, making it suitable for spatial planning.

\paragraph{First-person overlay.}
Using the current depth observation and camera geometry, we project the waypoint and corridor into the current RGB view. The projection is drawn over the observed image and passed through the navigator's visual stream. Unlike future-image generation, this interface preserves the real observation and adds only route guidance. It links the prediction to visible passages, doors, and obstacles in the current view.

Both interfaces contain the same short- and mid-horizon route information. The canvas emphasizes top-down geometry, whereas the overlay emphasizes egocentric visual grounding. Comparing them isolates the effect of visual reference frame without changing the future-state target.

\begin{table*}[t]
\centering
\caption{Comparison with prior methods on the R2R-CE and RxR-CE validation-unseen splits. Pano., Odo., Depth, and S.RGB denote panoramic observations, odometry, depth, and a single RGB camera, respectively. Reference indicates the publication venue. Results are taken from the corresponding papers. Methods marked with $^{*}$ use a learned waypoint predictor, and -- denotes an unreported result.}
\label{tab:main_results}
\resizebox{\textwidth}{!}{
\renewcommand{\arraystretch}{1.3}
\begin{tabular}{l c cccc cccc cccc}
\toprule
& & \multicolumn{4}{c}{Observation}
& \multicolumn{4}{c}{R2R-CE Val-Unseen}
& \multicolumn{4}{c}{RxR-CE Val-Unseen} \\
\cmidrule(lr){3-6}
\cmidrule(lr){7-10}
\cmidrule(lr){11-14}
Method
& Reference
& Pano.
& Odo.
& Depth
& S.RGB
& NE$\downarrow$
& OSR$\uparrow$
& SR$\uparrow$
& SPL$\uparrow$
& NE$\downarrow$
& SR$\uparrow$
& SPL$\uparrow$
& nDTW$\uparrow$ \\
\midrule

R2R-CMTP~\cite{chen2021topological}
& CVPR'21
& $\checkmark$ & $\checkmark$ & $\checkmark$ &
& 7.90 & 38.0 & 26.0 & 22.0
& -- & -- & -- & -- \\

HPN+DN$^{*}$~\cite{krantz2021waypoint}
& ICCV'21
& $\checkmark$ & $\checkmark$ & $\checkmark$ &
& 6.31 & 40.0 & 36.0 & 34.0
& -- & -- & -- & -- \\

CMA$^{*}$~\cite{hong2022bridging}
& CVPR'22
& $\checkmark$ & $\checkmark$ & $\checkmark$ &
& 6.20 & 52.0 & 41.0 & 36.0
& 8.76 & 26.5 & 22.1 & 47.0 \\

Sim2Sim$^{*}$~\cite{krantz2022sim}
& ECCV'22
& $\checkmark$ & $\checkmark$ & $\checkmark$ &
& 6.07 & 52.0 & 43.0 & 36.0
& 8.76 & 26.5 & 22.1 & -- \\

GridMM$^{*}$~\cite{wang2023gridmm}
& ICCV'23
& $\checkmark$ & $\checkmark$ & $\checkmark$ &
& 5.11 & 61.0 & 49.0 & 41.0
& -- & -- & -- & -- \\

HNR$^{*}$~\cite{wang2024lookahead}
& CVPR'24
& $\checkmark$ & $\checkmark$ & $\checkmark$ &
& 4.42 & 67.0 & 61.0 & 51.0
& 5.50 & 56.3 & 46.7 & 63.5 \\

ETPNav$^{*}$~\cite{an2024etpnav}
& TPAMI'24
& $\checkmark$ & $\checkmark$ & $\checkmark$ &
& 4.71 & 65.0 & 57.0 & 49.0
& 5.64 & 54.7 & 44.8 & 61.9 \\

Instruc-Nav~\cite{long2024instructnav}
& CoRL'24
& $\checkmark$ & $\checkmark$ & $\checkmark$ &
& 6.89 & -- & 31.0 & 24.0
& -- & -- & -- & -- \\

AO-Planner~\cite{chen2025affordances}
& AAAI'25
& $\checkmark$ & & $\checkmark$ &
& 5.55 & 59.0 & 47.0 & 33.0
& 7.06 & 43.3 & 30.5 & 50.1 \\

\midrule
Seq2Seq~\cite{krantz2020beyond}
& ECCV'20
& & & $\checkmark$ & $\checkmark$
& 7.77 & 37.0 & 25.0 & 22.0
& 12.10 & 13.9 & 11.9 & -- \\

LAW~\cite{raychaudhuri2021language}
& EMNLP'21
& & $\checkmark$ & $\checkmark$ & $\checkmark$
& 6.83 & 44.0 & 35.0 & 31.0
& 10.90 & 8.0 & 8.0 & 38.0 \\

CM2~\cite{georgakis2022cross}
& CVPR'22
& & $\checkmark$ & $\checkmark$ & $\checkmark$
& 7.02 & 41.0 & 34.0 & 27.0
& -- & -- & -- & -- \\

WS-MGMap~\cite{chen2022weakly}
& NeurIPS'22
& & $\checkmark$ & $\checkmark$ & $\checkmark$
& 6.28 & 47.0 & 38.0 & 34.0
& -- & -- & -- & -- \\

NaVid~\cite{zhang2024navid}
& RSS'24
& & & & $\checkmark$
& 5.47 & 49.1 & 37.4 & 35.9
& 8.41 & 23.8 & 21.2 & -- \\

Uni-NaVid~\cite{zhang2024uni}
& RSS'25
& & & & $\checkmark$
& 5.58 & 53.3 & 47.0 & 42.7
& 6.24 & 48.7 & 40.9 & -- \\

NavMorph~\cite{yao2025navmorph}
& ICCV'25
& & & $\checkmark$ & $\checkmark$
& 5.75 & 56.9 & 47.9 & 33.2
& 8.85 & 30.8 & 22.8 & 44.2 \\

MapNav~\cite{zhang2025mapnav}
& ACL'25
& & $\checkmark$ & $\checkmark$ & $\checkmark$
& 4.93 & 53.0 & 39.7 & 37.2
& 7.62 & 32.6 & 27.7 & 43.5 \\

NaVILA~\cite{cheng2024navila}
& RSS'25
& & & & $\checkmark$
& 5.22 & 62.5 & 54.0 & 49.0
& 6.77 & 49.3 & 44.0 & 58.8 \\

StreamVLN~\cite{wei2025streamvln}
& ICRA'26
& & & & $\checkmark$
& 4.98 & 64.2 & 56.9 & 51.9
& 6.22 & 52.9 & 46.0 & 61.9 \\

Goal2Pixel~\cite{bao2026goal2pixel}
& arXiv'26
& & & $\checkmark$ & $\checkmark$
& 4.85 & 59.9 & 54.1 & 52.5
& 7.50 & 43.8 & 40.4 & 61.1 \\

JanusVLN~\cite{zeng2025janusvln}
& ICLR'26
& & & & $\checkmark$
& \textbf{4.78} & 65.2 & 60.5 & \textbf{56.8}
& 6.06 & \textbf{56.2} & \textbf{47.5} & 62.1 \\

\midrule

\textbf{FS-Nav (ours)}
& --
& & $\checkmark$ & $\checkmark$ & $\checkmark$& 4.84 & \textbf{66.1} & \textbf{60.7} & 55.9&\textbf{6.03}& 55.1 & 46.5 & \textbf{62.7}\\

\bottomrule
\end{tabular}
}
\end{table*}

\subsection{Future-State-Guided Navigator}

The navigator is built on Qwen2.5-VL-3B-Instruct~\cite{bai2025qwen2} and receives the instruction, observation history, current observation, and future-state interface $Z_t^k$. For the canvas variant, the top-down canvas is appended as the final image. For the first-person variant, the overlaid RGB frame is supplied through the visual stream. The action space and supervision remain the same for both variants.

Following our VLN backbone, the navigator autoregressively generates a textual action sequence $A_t=(a_{t,1},\ldots,a_{t,L})$. The generated tokens are parsed into executable forward, turning, and stop commands. We train the navigator with supervised language modeling~\cite{zhang2024navid,wu2025event}:
\begin{equation}
    \mathcal{L}_{\mathrm{nav}}
    =-\sum_{l=1}^{L}
    \log p_{\theta}\!\left(
    a_{t,l}\mid a_{t,<l},O_{\leq t},\mathcal{L},Z_t^k
    \right).
\end{equation}
Only the target action tokens contribute to the loss; prompts and visual placeholders are used as conditioning context.

\begin{figure*}[t]
\centering
\includegraphics[width=\textwidth]{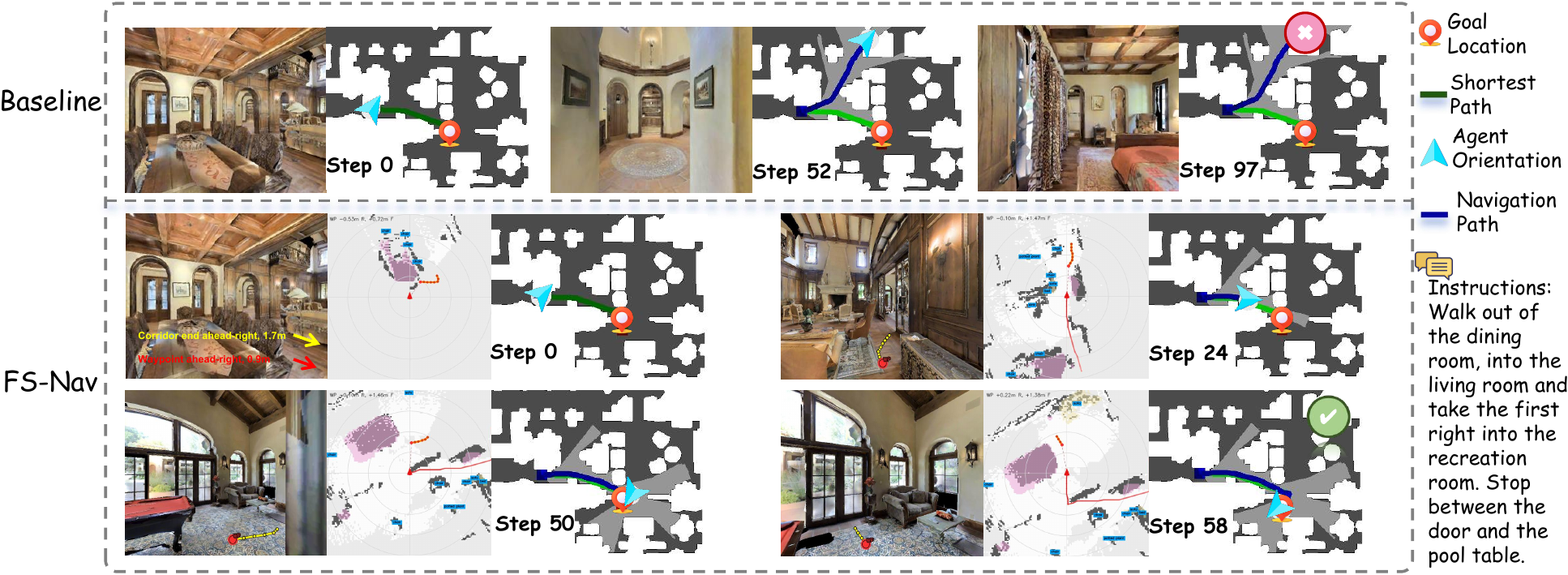}
\caption{Qualitative comparison on the same R2R-CE episode. The baseline deviates from the reference route and fails to reach the goal, whereas FS-Nav uses the predicted waypoint--corridor guidance, rendered through the first-person overlay and top-down canvas, to follow the instructed route and stop near the goal. The green and blue lines denote the shortest and executed navigation paths, respectively. When the predicted guidance lies outside the first-person field of view, an arrow indicates its direction.}
\label{fig:qualitative}
\end{figure*}

\section{Experiments}

\subsection{Experimental Setup}

\paragraph{Datasets and Evaluation Metrics.}
We evaluate FS-Nav on
VLN-CE benchmarks using the validation-unseen splits of R2R-CE~\cite{krantz2020beyond} and RxR-CE~\cite{ku2020room}.
Following prior works~\cite{zhang2024navid,zhu2026navida}, we report Navigation Error (NE), Oracle Success Rate (OSR), Success Rate (SR), Success weighted by Path Length (SPL) and normalized Dynamic Time Warping (nDTW).

\paragraph{Implementation Details.}
For future-state construction, we set the waypoint horizon to $d_{\mathrm{w}}=1.5$ meters and the corridor endpoint to $d_{\mathrm{c}}=3.0$ meters, with at most 20 future steps. These distances are chosen to favor visibility from the agent's current viewpoint: a shorter waypoint horizon often places the target below the camera's field of view, whereas corridor geometry beyond $3.0$ meters is more likely to be occluded or outside the visible range. We train the Future State World Model on R2R, RxR, and ScaleVLN training trajectories for 20 epochs using AdamW with a learning rate of $1\times10^{-4}$, a batch size of 16, and cosine decay. We set $\lambda_{\mathrm{w}}=0.5$, $\lambda_{\mathrm{dice}}=0.5$, $\lambda_{\mathrm{cl}}=0.5$, $\lambda_{\mathrm{tv}}=0.1$, $\lambda_{\mathrm{pts}}=1.0$, and $\lambda_{\mathrm{curv}}=0.1$. After training, the world model is frozen throughout navigator training. The navigator is initialized from Qwen2.5-VL-3B-Instruct and takes up to eight historical RGB observations together with the current observation and language instruction to autoregressively predict textual action chunks. We first train it for one epoch on R2R and RxR, and then use the resulting checkpoint to collect DAgger trajectories on their training splits. Finally, we continue training the same navigator on the combined R2R, RxR, ScaleVLN, and DAgger data to obtain the final model. Navigator training uses a learning rate of $2\times10^{-5}$, an effective batch size of 128, a linear schedule, bfloat16 precision, and DeepSpeed ZeRO-2. Unless otherwise specified, we use both future-state interfaces. Depth, odometry, and predicted semantics are used only to update the spatial map.

\subsection{Main Results on VLN Benchmarks}

\paragraph{Quantitative Results.}
Table~\ref{tab:main_results} compares FS-Nav with panoramic waypoint-based agents and non-panoramic VLN methods under their reported sensor configurations. On R2R-CE, FS-Nav achieves 60.7 SR and 55.9 SPL, obtaining the highest OSR and SR among non-panoramic methods while remaining competitive in path efficiency. On the more challenging RxR-CE benchmark, FS-Nav reaches 55.1 SR, 46.5 SPL, and 62.7 nDTW. Although JanusVLN attains slightly higher SR and SPL, FS-Nav achieves lower NE and higher nDTW, indicating more accurate final localization and stronger alignment with the reference route. Overall, FS-Nav performs competitively across both benchmarks and establishes strong results without panoramic observations.

\paragraph{Qualitative Results.}
Figure~\ref{fig:qualitative} compares the baseline and FS-Nav under the same instruction and initial observation. The baseline gradually deviates from the reference route and enters an incorrect region. In contrast, the predicted waypoint and corridor provide FS-Nav with explicit local targets and route-continuation guidance, allowing it to complete the required turns and stop near the goal in fewer steps.

\subsection{Ablation and Analysis}

We analyze FS-Nav from six aspects: future-state interfaces, waypoint and corridor, the contribution of future-state guidance, navigator training distribution, future-state prediction quality, and computational efficiency. For controlled and reproducible comparisons, we vary one factor at a time while keeping the remaining settings unchanged and use fixed random seeds throughout training and evaluation. Unless otherwise specified, all models are trained on the R2R training split and evaluated on the R2R-CE validation-unseen split.
\paragraph{Effect of Future-state interfaces.}
\begin{table}[t]
\centering
\caption{Ablation of future-state interfaces. All variants use predictions from the same frozen world model.}
\label{tab:interface_ablation}
\resizebox{\columnwidth}{!}{
\begin{tabular}{cc|ccccc}
\toprule
Canvas & Overlay & NE$\downarrow$ & SR$\uparrow$ & OSR$\uparrow$ & SPL$\uparrow$ & nDTW$\uparrow$ \\
\midrule
& & 6.85 & 40.9 & 50.1 & 36.6 & 52.9 \\
$\checkmark$ & & 6.48 & 42.9 & 55.6 & 37.3 & 53.7 \\
& $\checkmark$ & 6.25 & 43.6 & 53.1 & 38.8 & 55.5 \\
$\checkmark$ & $\checkmark$ & \textbf{6.14} & \textbf{44.4} & \textbf{56.2} & \textbf{39.6} & \textbf{56.1} \\
\bottomrule
\end{tabular}}
\vspace{-0.3cm}
\end{table}
We first examine which visual interface most effectively communicates the same predicted future state to the navigator and whether the two reference frames provide complementary guidance.
As shown in Table~\ref{tab:interface_ablation},  both interfaces improve over the no-interface baseline, confirming that the predicted future state is useful under either visual reference frame. The overlay performs better overall than the canvas (43.6 vs. 42.9 SR), suggesting that projecting the route into the current view facilitates visual grounding. Using both interfaces gives the best overall results, as they provide complementary guidance: the first-person overlay grounds the predicted route in the current egocentric view, while the top-down canvas provides broader spatial context beyond the camera's field of view.

\begin{table}[t]
\centering
\caption{Ablation of the future-state representation.}
\label{tab:state_ablation}
\resizebox{\columnwidth}{!}{
\begin{tabular}{cc|ccccc}
\toprule
Waypoint & Corridor & NE$\downarrow$ & SR$\uparrow$ & OSR$\uparrow$ & SPL$\uparrow$ & nDTW$\uparrow$ \\
\midrule
& & 6.85 & 40.9 & 50.1 & 36.6 & 52.9 \\
$\checkmark$ & & 6.32 & 43.0 & 55.0 & 38.2 & 54.7 \\
& $\checkmark$ & 6.23 & 43.7 & 55.6 & 38.9 & 55.4 \\
$\checkmark$ & $\checkmark$ & \textbf{6.14} & \textbf{44.4} & \textbf{56.2} & \textbf{39.6} & \textbf{56.1} \\
\bottomrule
\end{tabular}}
\vspace{-0.3cm}
\end{table}

\paragraph{Effect of waypoint and corridor.}
We examine whether the short-horizon waypoint and mid-horizon corridor provide complementary guidance. The waypoint-only variant specifies the next local target, while the corridor-only variant describes the upcoming route without explicitly marking an immediate target.
As shown in Table~\ref{tab:state_ablation}, both components improve navigation independently. The corridor achieves higher SR than the waypoint alone (43.7 vs.\ 43.0), showing the value of mid-horizon route geometry. Combining them gives the best performance, confirming their complementary roles: the waypoint anchors the next local target, while the corridor indicates how the route continues beyond it.

\begin{figure}[t]
\centering
\begin{subfigure}{\columnwidth}
    \centering
    \includegraphics[width=\columnwidth]{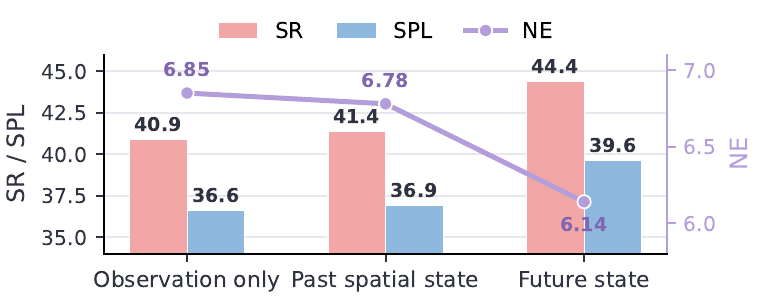}
    \caption{Future-state guidance versus accumulated past spatial context.}
    \label{fig:guidance_source}
\end{subfigure}

\vspace{1mm}

\begin{subfigure}{\columnwidth}
    \centering
    \includegraphics[width=\columnwidth]{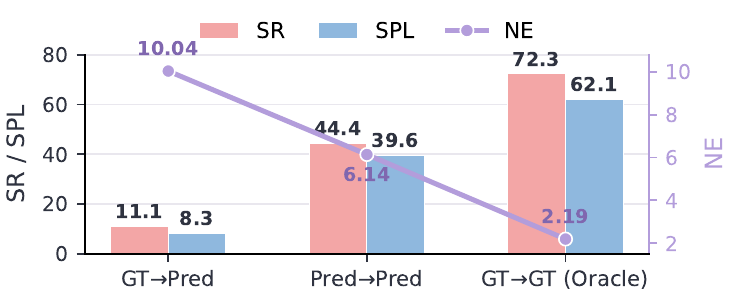}
    \caption{Effect of navigator training and inference distributions.}
    \label{fig:training_distribution}
\end{subfigure}
\caption{Ablation of future-state guidance and navigator training distribution on R2R-CE validation unseen. Bars report SR and SPL, while the red line reports NE. GT$\rightarrow$GT uses future states derived from the validation reference trajectory and serves only as an oracle upper bound.}
\label{fig:training_ablation}

\end{figure}
\begin{figure}[t]
\centering
\includegraphics[width=\columnwidth]{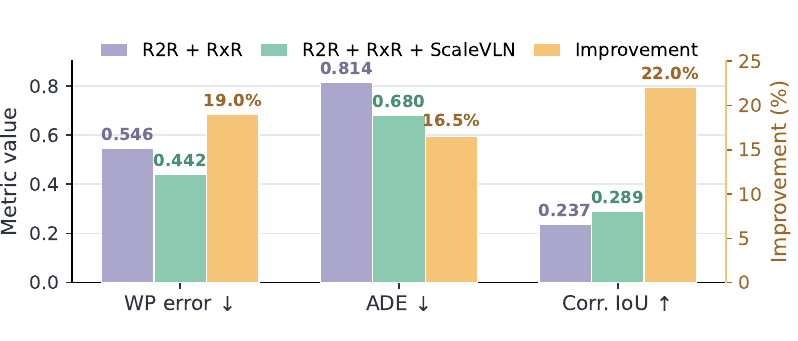}
  \caption{Future-state prediction quality on R2R validation unseen. WP error and ADE measure waypoint and ordered-path accuracy, respectively, while corridor IoU measures mask overlap. Improvement denotes relative
  error reduction or IoU increase.}
\label{fig:wm_quality}
\end{figure}

\paragraph{Future-state guidance and training distribution.}
We examine two questions: whether the gains arise from predicted future-route guidance rather than additional past spatial context, and whether the navigator should be trained with predicted future states. The observation-only baseline uses visual history alone. The past spatial state control additionally provides the accumulated spatial map~$M_t$ but removes waypoint and corridor predictions, while the future-state condition corresponds to the full FS-Nav model. GT$\rightarrow$Pred trains the navigator with ground-truth future states but evaluates it with predicted states. In contrast, Pred$\rightarrow$Pred uses outputs from the same frozen world model during both training and evaluation. GT$\rightarrow$GT uses ground-truth future states at both stages as a diagnostic oracle.

As shown in Fig.~\ref{fig:training_ablation} (a), adding only past spatial context improves SR by 0.5 points, whereas future-state guidance provides a further 3.0-point gain. This suggests that the improvement mainly comes from anticipated route information rather than additional spatial memory alone. Figure~\ref{fig:training_ablation} (b) further shows the importance of matching the training and inference distributions. GT$\rightarrow$Pred drops to 11.1 SR, indicating a severe mismatch when the navigator is trained with clean future states but receives imperfect predictions at test time. Pred$\rightarrow$Pred is exposed to such prediction errors during training and is therefore more robust at inference. GT$\rightarrow$GT achieves 72.3 SR, showing that accurate waypoint--corridor states can provide highly informative guidance.
The gap to Pred$\rightarrow$Pred suggests that future-state prediction quality remains a major bottleneck.

\paragraph{Future-state prediction quality.}\
We evaluate the frozen world model independently on R2R validation-unseen trajectories. WP error measures waypoint localization, ADE measures the average point-wise distance between the predicted and reference corridor polylines, and corridor IoU measures mask overlap. We compare world models trained on R2R and RxR with the same model additionally trained on ScaleVLN.

Figure~\ref{fig:wm_quality} shows consistent improvements across waypoint, ordered-path, and corridor-mask accuracy after adding ScaleVLN. In particular, ADE decreases by 16.5\%, confirming that broader trajectory supervision improves not only mask overlap but also the predicted route structure. We use this world model for all downstream FS-Nav experiments.

\paragraph{Efficiency Analysis.}
We measure world-model inference latency and deterministic interface-rendering latency with batch size one on the same hardware. The canvas and first-person overlay share the same future-state prediction, so their combination invokes the world model only once and adds only the cost of rendering.

\begin{figure}[t]
\centering
\includegraphics[width=1.0\columnwidth]{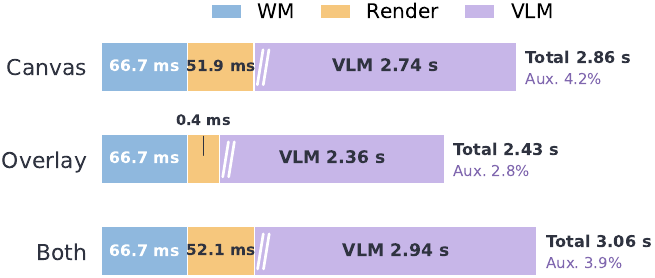}
\caption{Computation and interface cost on a single RTX 4090 with batch size one. All configurations share the same 9.45M-parameter Future State World Model and invoke it once per navigation step. The overlay rendering segment is enlarged for visibility and is not drawn to scale.}
\label{fig:efficiency}
\end{figure}

As shown in Figure~\ref{fig:efficiency}, future-state prediction takes 66.7ms across all configurations. Rendering costs 0.4ms for the first-person interface, 51.9ms for the canvas, and 52.1ms for their combination, while the corresponding VLM inference latencies are 2.36s, 2.74s, and 2.94s.
These results show that FS-Nav introduces limited computational overhead while providing explicit future-route guidance.

\section{Conclusion}
In this work, we explored a cognitively inspired perspective on navigation world modeling that organizes the anticipated future around goal-directed spatial states and route structure. We instantiated this perspective with FS-Nav, which predicts a short-horizon waypoint and a mid-horizon corridor from spatial memory and language instructions and communicates them to a VLM navigator through complementary visual interfaces. Experiments on standard VLN benchmarks demonstrate the effectiveness of this compact representation, highlighting navigation-relevant future-state modeling as a promising direction for VLN.

\bibliography{aaai2027}
\end{document}